\documentclass[final]{cvpr}

\usepackage{times}
\usepackage{epsfig}
\usepackage{graphicx}
\usepackage{amsmath}
\usepackage{amssymb}
\usepackage{subfigure}

\usepackage[pagebackref=true,breaklinks=true,colorlinks,bookmarks=false]{hyperref}

\def\cvprPaperID{8801} 
\def\confYear{CVPR 2021}
\begin{document}

\title{Generalizing to the Open World: Deep Visual Odometry with Online Adaptation}

\author{Shunkai Li \qquad
	Xin Wu \qquad
	Yingdian Cao \qquad
	Hongbin Zha\\
	Key Laboratory of Machine Perception (MOE), School of EECS, Peking University\\
	PKU-SenseTime Machine Vision Joint Lab\\
	{\tt\small \{lishunkai, wuxin1998, yingdianc\}@pku.edu.cn \quad zha@cis.pku.edu.cn} 
}

\maketitle

\begin{abstract}
	\hyphenpenalty=5000
	\tolerance=1000
	Despite learning-based visual odometry (VO) has shown impressive results in recent years, the pretrained networks may easily collapse in unseen environments. The large domain gap between training and testing data makes them difficult to generalize to new scenes. In this paper, we propose an online adaptation framework for deep VO with the assistance of scene-agnostic geometric computations and Bayesian inference. In contrast to learning-based pose estimation, our method solves pose from optical flow and depth while the single-view depth estimation is continuously improved with new observations by online learned uncertainties. Meanwhile, an online learned photometric uncertainty is used for further depth and pose optimization by a differentiable Gauss-Newton layer. Our method enables fast adaptation of deep VO networks to unseen environments in a self-supervised manner. Extensive experiments including Cityscapes to KITTI and outdoor KITTI to indoor TUM demonstrate that our method achieves state-of-the-art generalization ability among self-supervised VO methods.
	
\end{abstract}

\section{Introduction}
	\hyphenpenalty=5000
	\tolerance=1000
Estimating camera motion from monocular videos plays an essential role in many real-world applications, such as autonomous driving and robotics. This problem is usually solved by visual odometry (VO) or simultaneous localization and mapping (SLAM). Classic SLAM/VO methods~\cite{DSO,LSD,svo,orb} perform well in favorable conditions but often fail in challenging situations (\textit{e.g.} textureless region, dynamic object) due to the reliance on low-level features and hand-crafted pipeline. Since deep neural networks are able to extract high-level features and infer end-to-end by learning from data, many learning-based VO methods~\cite{onlinevo,savo,GeoNet,SfMLearner} have been proposed to break through the limitations of classic SLAM/VO. Among them, self-supervsied VO methods are able to jointly learn camera pose, depth and optical flow by minimizing photometric error~\cite{GeoNet}, which have shown promising results in recent years.

\begin{figure}
	\centering
	\subfigbottomskip=1pt 
	\subfigure{\includegraphics[width=0.96\linewidth]{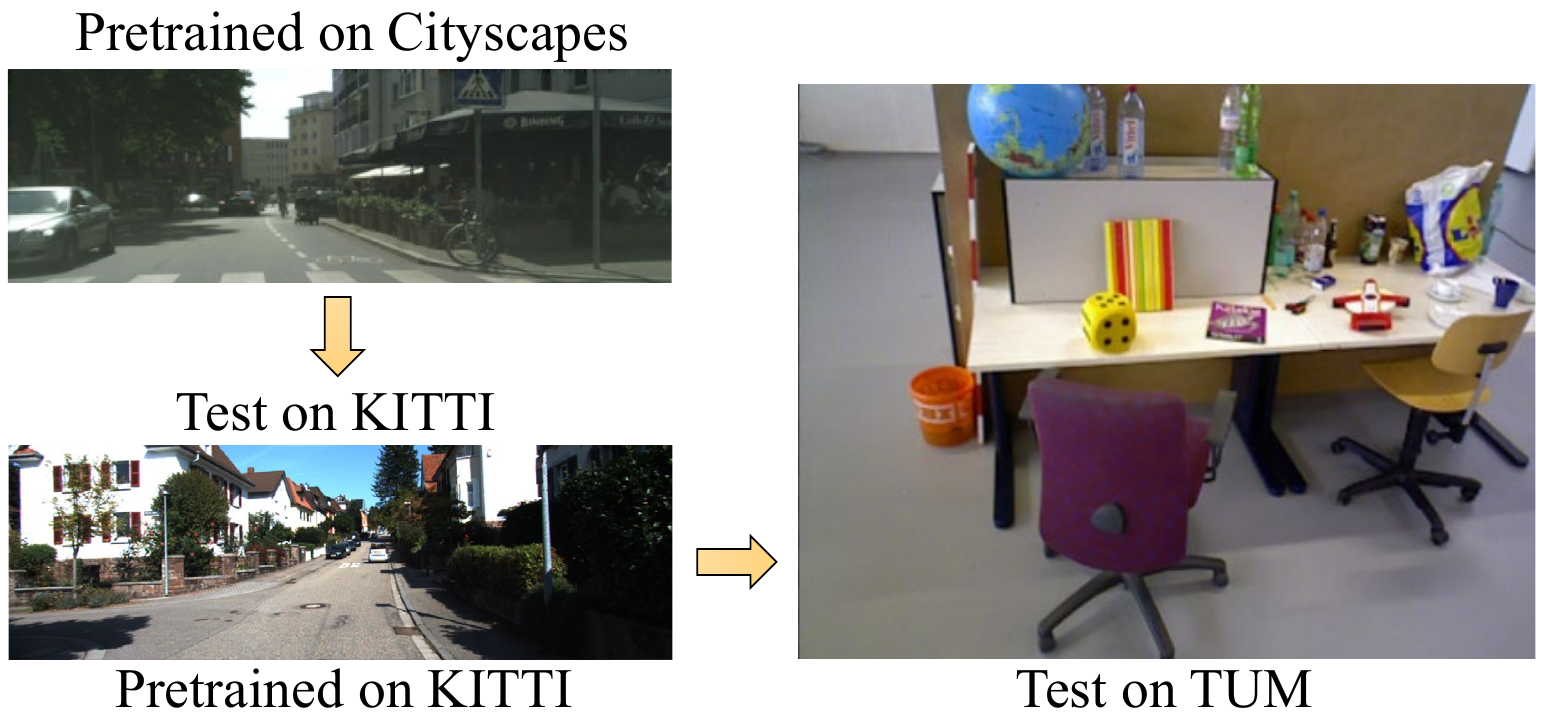}}  
	\subfigure{\includegraphics[height=4.5cm]{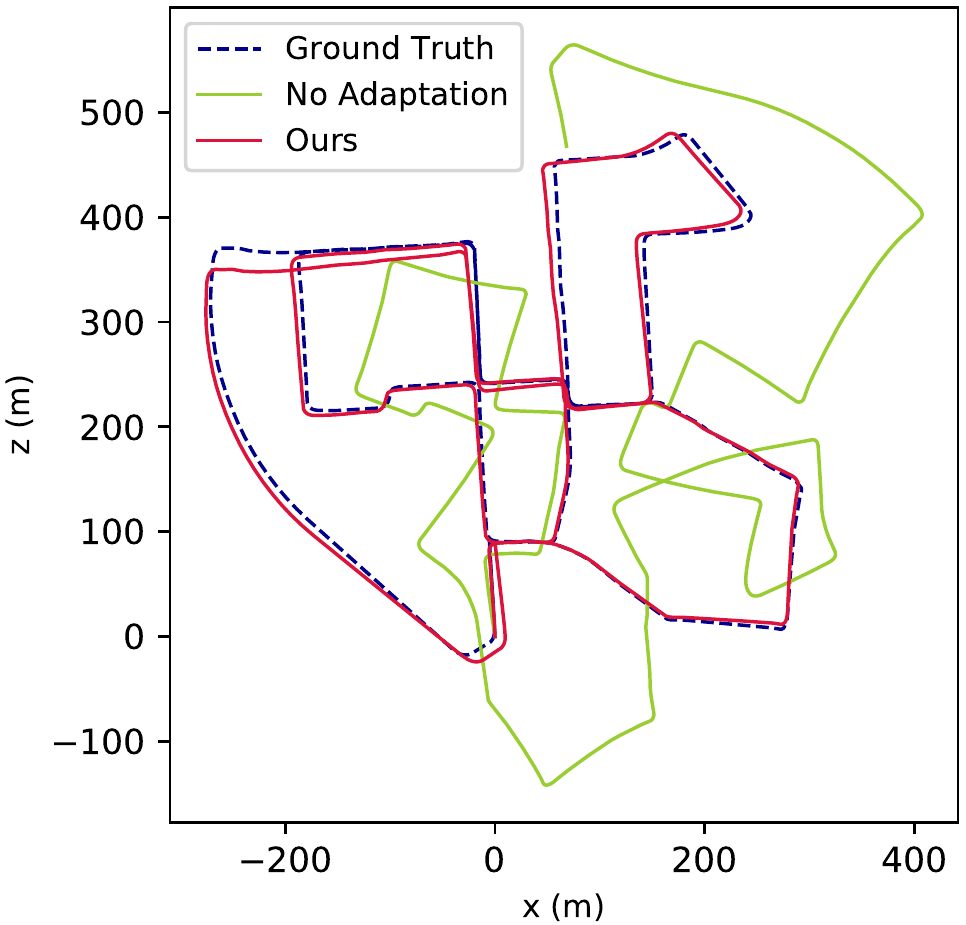}}
	\subfigure{\includegraphics[height=4.5cm]{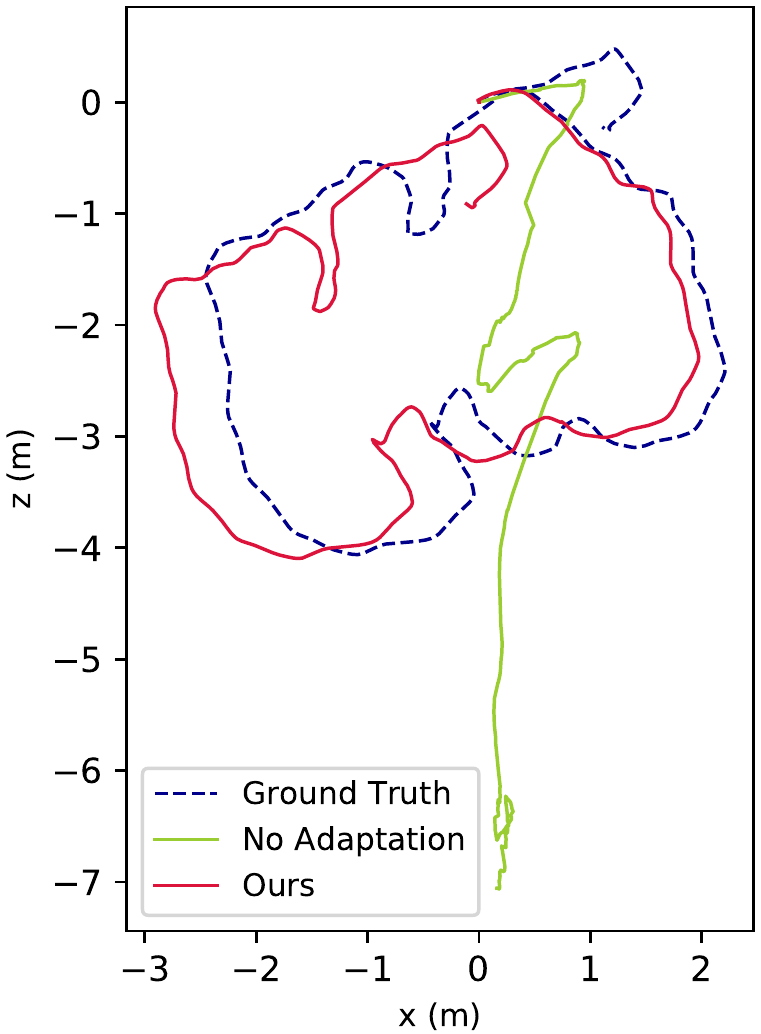}} 
	\caption{In this paper, we focus on the generalization ability to unseen environments of deep VO. When the test data are different from the training data, previous methods fail to generalize while our method still performs well with very small trajectory error.}
\end{figure}

However, learning-based VO often fails during inference when the scenes are different from the training data. The inability of pretrained VO to generalize to unseen environments limits its wide applications~\cite{onlinevo,open}. To this end, the pretrained networks are required to achieve real-time online adaptation in a self-supervised manner.

As a result, several previous works~\cite{onlinedepth,onlinevo,open} have been proposed to mitigate the domain generalization problem of stereo matching and VO. However, the performance is still much inferior to classic methods in terms of accuracy and the pretrained networks suffer from slow convergence. These methods treat VO as a black-box by learning all components (pose, depth, optical flow, etc.) but ignore well-defined geometric computations and optimization methods, which leads to slow convergence during online adaptation.

Existing deep VO methods predict depth by single-view estimation, which is an ill-posed problem~\cite{savo}. The learned depth has a strong reliance on the training dataset. During inference, the camera intrinsics, scene layouts and distances are usually different. Meanwhile, the camera pose is learned rather than calculated analytically, which requires favorable camera motion with sufficient disparity (\textit{e.g.} KITTI dataset). Therefore, these methods tend to fail when faced with unseen or more complicated motion patterns. In addition, existing learning-based methods do not explicitly ensure multi-view geometric consistency during inference, which leads to large scale drift in trajectories.

In order to improve the online adaptation of VO to unseen environments, we propose a self-supervised framework that combines the advantage of deep learning and geometric computations. The proposed framework utilizes scene-agnostic 3D geometry constraints and Bayesian inference formulations to speed up online adaptation. During inference, the single-view depth estimation is used as a prior of the current scene geometry and is continuously improved with incoming observations by a probabilistic Bayesian updating framework. The refined depth is used as Maximum A Posteriori (MAP) to train DepthNet for better estimation at the next timestep. Instead of predicting pose by PoseNet, our framework solves pose analytically from optical flow and refined depth. Meanwhile, in order to deal with observation noise, the proposed method online learns depth and photometric uncertainties which are used in the depth refinement process and differentiable Gauss-Newton optimization, respectively. Finally, the optimized pose, depth and flow are used for online self-supervision. Our framework ensures scale consistency by exploiting multi-view geometric constraints. The well-defined \textit{scene-agnostic} computation helps our VO framework achieve good generalization ability across different scene conditions. Our contributions can be summarized as follows:

\begin{itemize}
	\item We propose a generalizable deep VO that uses scene-agnostic geometric formulation and Bayesian inference to speed up self-supervised online adaptation.
	\item The predicted depth is continuously refined by a Bayesian fusion framework, which is further used to train depth and optical flow during online learning.
	\item We introduce online learned depth and photometric uncertainties for better depth refinement and differentiable Gauss-Newton optimization.
\end{itemize}

Our method achieves much better generalization than state-of-the-art baselines when tested cross different domains, including Cityscapes~\cite{cityscapes} to KITTI~\cite{kitti} and outdoor KITTI to indoor TUM~\cite{TUM} datasets. Meanwhile, we also achieve state-of-the-art depth estimation results on KITTI and NYUv2~\cite{nyu} datasets.


\section{Related works}
{\bf Learning-based VO} has been widely studied in recent years and shown impressive results~\cite{deepvo,beyond,deeptam}. DeepTAM~\cite{deeptam} mimics the framework of parallel tracking and mapping in classic SLAM/VO by using two networks for depth and pose estimation simultaneously. Xue~\textit{et al}~\cite{beyond} extends the VO pipeline to tracking, selecting memory and refining modules, which shows superior performance under challenging conditions. However, these methods require ground truth which is often impractical to obtain. In order to alleviate the need of ground truth data, self-supervised VO has been proposed. SfMLearner~\cite{SfMLearner} learns depth and pose simultaneously by minimizing photometric loss between warped and input image. Zhao \etal~\cite{towards} and Ranjan \etal~\cite{competitive} extend this idea to joint estimation of pose, depth and optical flow. Monodepth2~\cite{monodepth2} explicitly handles non-rigid and occluded cases which are against static-scene assumption. SAVO~\cite{savo} exploits spatial-temporal correlations over long sequence and utilizes RNN to reduce scale drift. In this paper, we use the depth network of Monodepth2~\cite{monodepth2} for single-view depth estimation.

{\bf Online adaptation} Most machine learning algorithms assume that the training and testing data are sampled from the same feature distribution. However, when the test data are different from the training set, most pretrained models suffer from a significant reduce in performance. In this situation, online learning~\cite{domainshift,lifelong} is an effective method to solve the domain shift problem. Previous methods use online gradient update~\cite{onlineSGD} and probabilistic filtering~\cite{stream} to accelerate domain adaptaion. In the computer vision field, Zhong \etal~\cite{open} proposes a self-supervised framework for stereo matching in the open world. Li \etal~\cite{onlinevo} proposes an online meta-learning algorithm for VO to continuously adapt to unseen environments. However, these methods learn all components by deep networks, leading to slow convergence and inferior performance. In contrast, our method combines the advantage of deep learning and well-defined geometric computations to achieve better generalization.

\begin{figure*}
	\centering
	\includegraphics[width=1\linewidth]{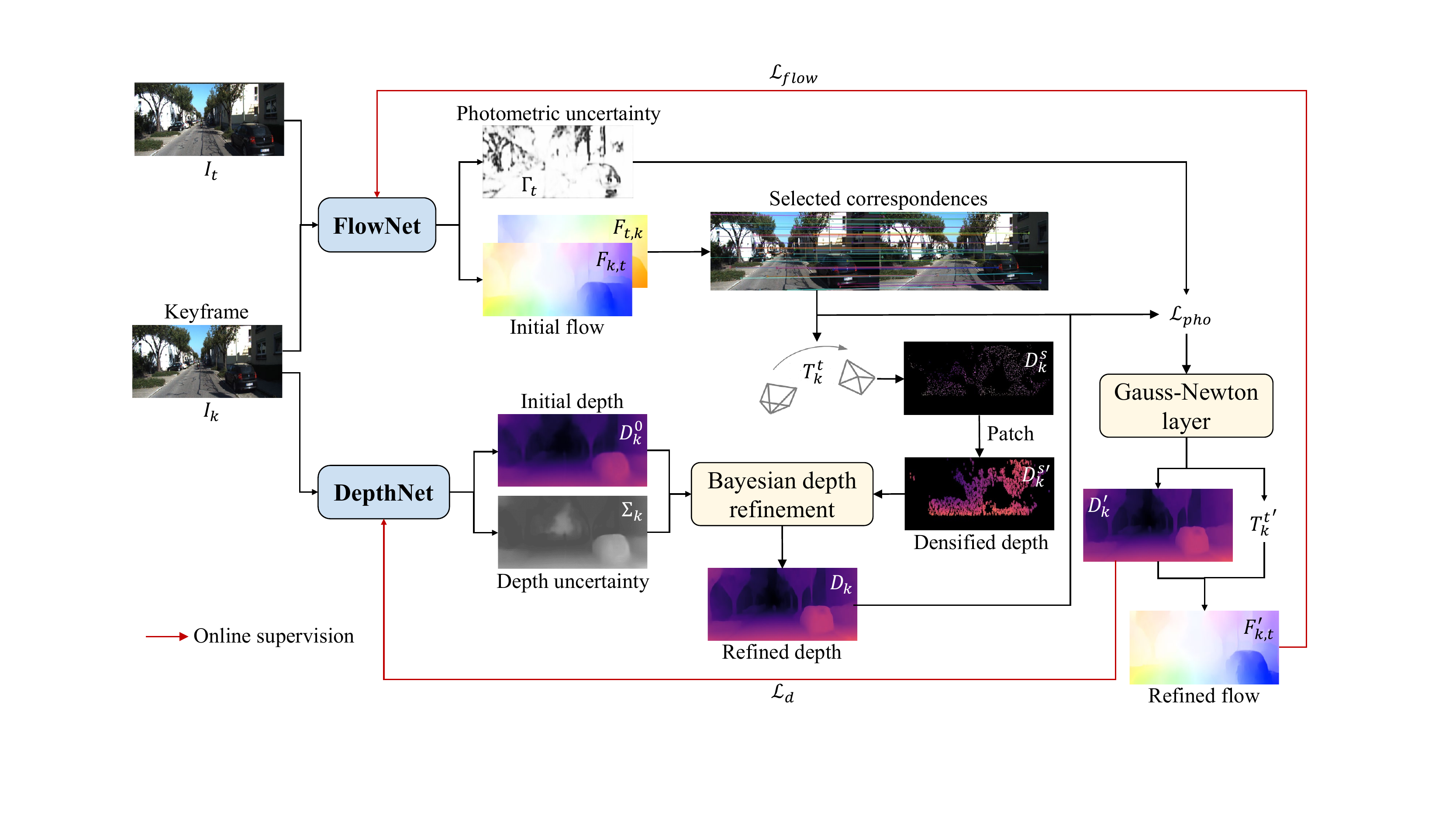}
	\caption{The framework of our online adaptation method. FlowNet predicts dense optical flow $F_{k,t},F_{t,k}$ and photometric uncertainty $\Gamma_{t}$, while DepthNet provides a prior depth estimation of the keyframe by estimating initial depth $D^{0}_{k}$ and uncertainty $\Sigma_{k}$. The relative pose $T^{t}_{k}$ is solved analytically from selected correspondences. During online adaptation, the initial depth $D^{0}_{k}$ is continuously improved with new triangulated depth patches in a Bayesian updating framework. The photometric loss weighted by $\Gamma_{t}$ is minimized by a differentiable Gauss-Newton layer. Finally, the optimized depth and pose are then used to self-supervise the online learning of DepthNet and FlowNet.}
	\label{fig:overview}
\end{figure*}

{\bf 3D Geometric computations} In classic 3D computer vision, the relative pose between two images and scene depth can be solved analytically by multi-view geometric constraints. Given a set of correspondences, the pose can be solved by epipolar geometry~\cite{epi,sampson} with 2D-2D matching or Perspective-n-Point (PnP)~\cite{pnp} with 3D-2D matching. The depth of each correspondence can be recovered by mid-point triangulation~\cite{orb}. On the other hand, the depth and pose can also be solved by minimizing photometric error~\cite{DSO,LSD} via classic optimizations. If more observations are available, the 3D map can be further refined by Bundle Adjustment (BA)~\cite{orb} or filtering~\cite{svo}. In this paper, we adopt a Bayesian depth fusion method to refine single-view depth estimation and propose a differentiable Gauss-Newton layer to minimize weighted photometric residuals.


\section{Method}

In this section, we will introduce our framework in detail. The system overview is illustrated in Fig.~\ref{fig:overview}. Firstly, the FlowNet predicts dense optical flow between the keyframe $I_{k}$ and current frame $I_{t}$ (Section~\ref{sec:flow}), and predicts photometric uncertainty map $\Gamma_{t}$ (Section~\ref{sec:residual}) as a side output. Meanwhile, the DepthNet estimates depth mean $D^{0}_{k}$ and uncertainty $\Sigma_{k}$ of keyframe, providing a prior of the current scene geometry (Section~\ref{sec:depth}). The relative pose $T^{t}_{k}$ is solved by essential matrix or PnP from selected flow correspondences. During online adaptation, we firstly reconstruct the sparse depth of $I_{k}$ by a differentiable triangulation module. Then, the prior keyframe depth $D^{0}_{k}$ is continuously improved by subsequent depth estimations in a Bayesian updating framework (Section~\ref{sec:bayesian}). Next, the differentiable Gauss-Newton layer minimizes the photometric loss of $I_{t}$ and warped image $\hat{I}_{t}$ weighted by predicted $\Gamma_{t}$ (Section~\ref{sec:GN}). Finally, the optimized depth $D^{'}_{k}$ and flow $F^{'}_{k,t}$ are used as pseudo ground truth to supervise the online learning of DepthNet and FlowNet (Section~\ref{sec:loss}).

\subsection{Pose recovery from optical flow}\label{sec:flow}
We use RAFT~\cite{raft} to learn dense optical flow $F_{k,t}$ between keyframe $I_{k}$ and current frame $I_{t}$. The optical flow between $I_{k}$ and $I_{t-1}$ is used as a prior to initialize current flow prediction. However, the predicted flow is not accurate for all pixels and the pose estimation error will increase if the displacement becomes small. Thus we select robust correspondences $(p_{k},p_{t})$ with good forward-backward flow consistency and moderate flow magnitude~\cite{dfvo}:
\begin{equation}
\|F_{k,t}(p_{k})+F_{t,k}(p_{t})\|<\delta_{1}, \quad \|F_{k,t}(p_{k})\|>\delta_{2},
\end{equation}
where we set $\delta_{1}=0.1, \delta_{2}=3$. We select $I_{t}$ as a new keyframe if the mean flow of robust correspondences is larger than 30. Benefiting from this keyframe-based scheme, the motion disparity between two frames are increased, enabling more accurate pose and depth estimation.

Given 2D correspondences between $p_{k},p_{t}$, the relative pose $T^{t}_{k}=[R|t]$ is computed by solving essential matrix $E$ with RANSAC~\cite{ransac} algorithm:

\begin{equation}
p^{T}_{t}K^{-T}EK^{-1}p_{k}=0, \quad E=[t]_{\times}R,
\label{eq:emat}
\end{equation}
where $K$ denotes camera intrinsics. The scale of up-to-scale pose $T^{t}_{k}$ is recovered by aligning triangulated sparse depth (detailed in Section~\ref{sec:bayesian}) with keyframe depth. However, when confronted with small translation or pure rotation, the 2D-2D estimation fails. In these cases, we recover pose with PnP~\cite{pnp} by minimizing reprojection error:

\begin{equation}
e_{r}=\sum ||KT^{t}_{k}D_{k}K^{-1}p_{k} - p_{t}||_{2},
\label{eq:pnp}
\end{equation}
where the 2D correspondences in $I_{k}$ are lifted to 3D with depth $D_{k}$ (detailed in Section~\ref{sec:depth}-\ref{sec:bayesian}) and intrinsics $K$.


\subsection{Depth modeling}\label{sec:depth}
In this paper, we model the depth estimation and updating in a unified Bayesian framework. The inverse depth $z_{i}=\frac{1}{d_{i}}$ of every pixel $i$ is used since it obeys Gaussian-like distribution and is more robust to distant objects. For inverse depth measurement $z^{t}_{i}$ at time $t$, we model the good measurement as Gaussian distribution around the ground truth $z_{i}$ while the bad one is regarded as observation noise which is uniformly distributed within the interval $[z^{\text{min}}_{i}, z^{\text{max}}_{i}]$. For every new observation $z^{t}_{i}$, the probability of being a inlier is $\rho^{t}_{i}$. Thus $z^{t}_{i}$ is modeled as~\cite{svo}:
\begin{equation}
p(z^{t}_{i}|z_{i},\rho^{t}_{i}):=\rho^{t}_{i}\mathcal{N}(z^{t}_{i}|z_{i},\tau^{2}_{i})+(1-\rho^{t}_{i})\mathcal{U}(z^{t}_{i}|z^{\text{min}}_{i},z^{\text{max}}_{i}),
\end{equation}
where $\tau^{2}_{i}$ denotes the variance of a good measurement. We follow~\cite{svo} to set inverse depth variance $\tau^{2}_{i}$ as the photometric disparity error of one pixel.

During online inference, we seek to find the Maximum A Posteriori (MAP) estimation of $z^{t}_{i}$ at each timestep, which can be approximated~\cite{beta} by the product of a Gaussian distribution for $z^{t}_{i}$ and a Beta distribution for inlier ratio $\rho^{t}_{i}$:
\begin{equation}\label{eq:update}
q(z^{t}_{i},\rho^{t}_{i}|a^{t}_{i},b^{t}_{i},\mu^{t}_{i},{\sigma^{t}_{i}}^{2}):=Beta(\rho^{t}_{i}|a^{t}_{i},b^{t}_{i})\mathcal{N}(z^{t}_{i}|\mu^{t}_{i},{\sigma^{t}_{i}}^{2}),
\end{equation}
where $a^{t}_{i},b^{t}_{i}$ are the parameters in Beta distribution, and $\mu^{t}_{i}, {\sigma^{t}_{i}}^{2}$ the mean and variance of Gaussian depth estimate.

The depth of \textit{keyframe} is initialized with single-view estimation $d^{0}_{k}\in D^{0}_{k}$ and inverse depth uncertainty $\sigma^{0}_{i}\in\Sigma_{k}$ from DepthNet as follows:
\begin{equation}
\begin{aligned}
	&\mu^{0}_{i}=\frac{1}{d^{0}_{k}}, \quad \sigma^{0}_{i}\in\Sigma_{k}, \quad z^{\text{max}}_{i}=\mu^{0}_{i}+\sigma^{0}_{i}, \\
	&z^{\text{min}}_{i}=
	\left\{  
	\begin{array}{lr}  
	\mu^{0}_{i}-\sigma^{0}_{i}, \qquad \text{if  } \mu^{0}_{i}-\sigma^{0}_{i}>0  \\  
	1e^{-6}, \indent\qquad \text{else} \\  
	\end{array}  
	\right.  
\end{aligned}\label{eq:init}
\end{equation}
During adaptation, the DepthNet online learns the prior knowledge of the new scene geometry. Besides, the learned uncertainties can also serve to gauge the reliability in probabilistic depth fusion.


\subsection{Online depth refinement}\label{sec:bayesian}
Given the relative pose $T^{t}_{k}$ and 2D correspondences, the subsequent depth estimation of keyframe can be further calculated by two-view triangulation~\cite{orb}:
\begin{equation}
    d^{k}_{i}=\mathop{\arg\min}_{d^{k}_{i}}[\text{dis}(L_{k},d^{k}_{i})^{2}+\text{dis}(L_{t},d^{k}_{i})]^{2},
\end{equation}
where dis() denotes the distance between $d^{k}_{i}$ and two camera rays $L_{k}, L_{t}$ generated from 2D correspondences. The mid-point triangulation is naturally differentiable, enabling our VO framework to perform end-to-end online learning.

\begin{figure}
	\centering
	\includegraphics[width=0.98\linewidth]{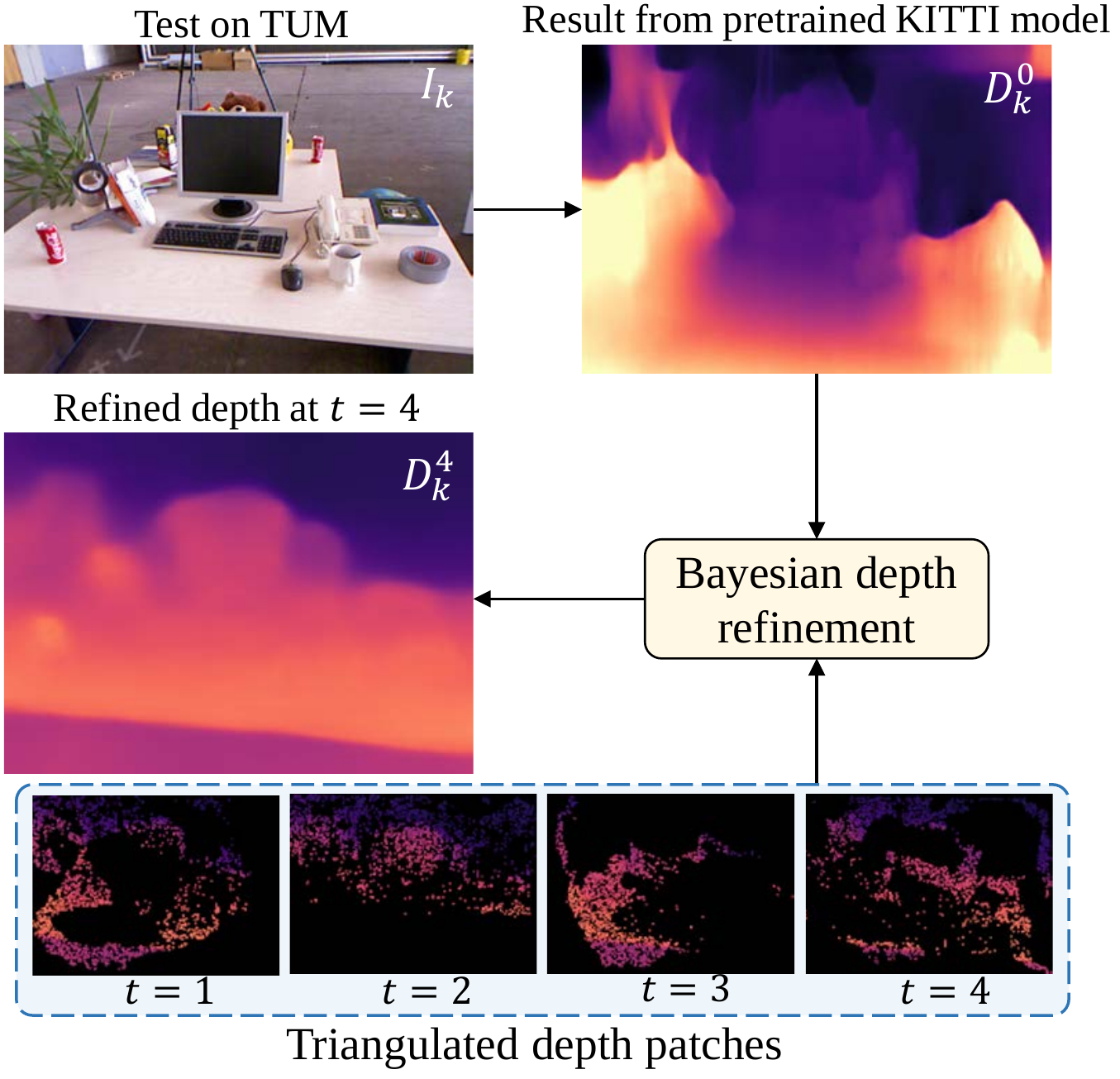}
	\caption{Illustration of online depth refinement process. When tested on TUM dataset, the pretrained network predicts erroneous depth. The initial guess is continuously updated by triangulated depth patches in a Bayesian refinement framework which becomes much more accurate after only 4 timesteps.}
	\label{fig:depth_refinement}
\end{figure}

The triangulated depth map $D^{s}_{k}$ is usually very sparse ($\sim$2000 points) and we densify each point with a local $3\times3$ patch ${D^{s}_{k}}^{'}$. The depth of each patch pixel is assumed the same as the central point. The patch-based representation allows larger region of depth filtering and provides more valid gradients with a wider basin of convergence.

During online adaptation, ${D^{s}_{k}}^{'}$ is used to update the prior depth estimate to get a MAP estimation $z^{t}_{i}$ according to Eq.~\ref{eq:update} as illustrated in Fig.~\ref{fig:depth_refinement}. Meanwhile, the parameters $a^{t}_{i},b^{t}_{i}, \mu^{t}_{i}, {\sigma^{t}_{i}}^{2}$ in Eq.~\ref{eq:update} are incrementally updated by Bayesian formulation. The updating method can be found in the supplementary materials. We assume the inverse depth $z^{t}_{i}$ have converged to the ground truth $z_{i}$ once the uncertainty ${\sigma^{t}_{i}}^{2}$ is lower than a threshold.


\subsection{Photometric residuals with learned uncertainty}\label{sec:residual}
Given the estimated pose $T^{t}_{k}$ and refined depth $D_{k}$, one can synthesize $\hat{I}_{t}$ by warping $I_{k}$ to the target image $I_{t}$~\cite{SfMLearner}:
\begin{equation}
	p_{t}\sim KT^{t}_{k}D_{k}(p_{k})K^{-1}p_{k},
\end{equation}

However, view synthesis builds on the photometric constancy assumption, which is often violated in practice. In order to alleviate this issue, we regard these corner cases as \textit{observation noise} and use deep neural network to predict a posterior probability distribution $p(I|\mu_{I},\gamma)$ for each RGB pixel parametrized by mean $\mu_{I}$ and variance $\gamma\in\Gamma$ over ground truth intensity $I$. By assuming the observation noise to be Laplacian, the online learning process can be formulated as minimizing the negative log-likelihood, which can be converted to a weighted photometric loss:
\begin{equation}\label{eq:photo}
\mathcal{L}_{pho}=\sum-\text{log }p(I|\mu_{I},\gamma)=\frac{\|\hat{I}_{t}-I_{t}\|_{1}}{\Gamma_{t}}+\text{log}\Gamma_{t},
\end{equation}
where $\Gamma_{t}$ denotes photometric uncertainty map.


\subsection{Differentiable Gauss-Newton optimization}\label{sec:GN}
Furthermore, we propose to use a differentiable Gauss-Newton~\cite{DSO} layer to miminize $\mathcal{L}_{pho}$ for optimized depth $D^{'}_{k}$ and pose ${T^{t}_{k}}^{'}$. The predicted $\Gamma_{t}$ in Eq.~\ref{eq:photo} improves the robustness to illumination change and occlusions. Specifically, starting with an initial depth and pose $D_{k},T^{t}_{k}$, we compute the weighted photometric loss $r_{i}(p)$ for each pixel $p_{i}$ in all frames $I_{i}$ among two keyframes $I_{k_{1}}, I_{k_{2}}$:
\begin{equation}
	r_{i}(p)=\frac{\hat{I}_{i}(p_{i})-I_{i}(p)}{\gamma_{i}}, \quad \gamma_{i}\in\Gamma_{t}
\end{equation}
The first order derivatives with respect to $D_{k}$ and $T^{t}_{k}$ are:
\begin{equation}
	J^{D}_{i}(p)=\frac{1}{\gamma_{i}}\frac{\partial \hat{I}_{i}(p_{i})}{\partial p_{i}}\frac{\partial p_{i}}{\partial D_{k}(p)},~~~
	J^{T}_{i}(p)=\frac{1}{\gamma_{i}}\frac{\partial \hat{I}_{i}(p_{i})}{\partial p_{i}}\frac{\partial p_{i}}{\partial T^{t}_{k}}
\end{equation}
Thus the increment $\delta$ to the current estimation is:
\begin{equation}
\delta=-(J^{T}J)^{-1}J^{T}r, \quad J=[J^{D}\ J^{T}]
\end{equation}
where $J$ denotes the stack of Jacobians $\{J_{i}(p)\}$ and $r$ denotes the stack of weighted photometric residuals $\{r_{i}(p)\}$. The Gauss-Newton algorithm is naturally differentiable and we implement it as a layer in neural network. In practice, we find that it converges within only 3 iterations.


\subsection{Loss functions}\label{sec:loss}
We propose to use the following loss functions to online learn DepthNet and FlowNet in a self-supervised manner.

{\bf Smoothness loss} We introduce an edge-aware loss for depth and flow to enforce local smoothness:
\begin{equation}
\begin{aligned}
\mathcal{L}_{smooth}(G)=\frac{1}{N}\sum_{x,y}&\|\nabla_{x}G(x,y)\|e^{-\|\nabla_{x}I(x,y)\|}+ \\&\|\nabla_{y}G(x,y)\|e^{-\|\nabla_{y}I(x,y)\|},
\end{aligned}
\end{equation}
where $G$ denotes optical flow or depth.

{\bf Depth loss} We derive a loss function of depth by evaluating the negative log-likelihood of the estimated inverse depth $\tilde{z}_{i}$ with uncertainty $\sigma_{i}$ defined in Eq.~\ref{eq:init}. This allows the network to atenuate the cost of difficult regions and to focus more on well explained parts. We assume a Laplacian distribution of inverse depth residuals:
\begin{equation}
p(\tilde{z}_{i}|\mu_{i},\sigma_{i})=\frac{1}{2\sigma_{i}}\text{exp}\left(-\frac{|\tilde{z}_{i}-\mu_{i}|}{\sigma_{i}}\right)
\end{equation}
We use refined inverse depth $z^{'}_{k}$ as $\mu_{i}$ for self-supervision. Thus the negative log-likelihood becomes:
\begin{equation}\label{eq:depthloss}
\mathcal{L}_{d}=\sum-\text{log }	p(\tilde{z}_{i}|\mu_{i},\sigma_{i})=\frac{\|1/D^{'}_{k}-1/D_{k}\|_{1}}{\Sigma_{k}}+\text{log}\Sigma_{k}
\end{equation}
Intuitively, the network will tune the depth uncertainty $\sigma_{i}$ that best minimize the depth loss $\|1/D^{'}_{k}-1/D_{k}\|_{1}$ while being subject to the regularization term $\text{log}\Sigma_{k}$. In order to enforce depth continuity, we modify Eq.~\ref{eq:depthloss} to:
\begin{equation}
\mathcal{L}_{d}=\frac{\|1/D^{'}_{k}-1/D_{k}\|_{1}}{\Sigma_{k}}+\text{log}\Sigma_{k}+\mathcal{L}_{smooth}(D_{k})
\end{equation}

{\bf Flow loss} The optimized depth and pose $D^{'}_{k},{T^{t}_{k}}^{'}$ can be used to synthesize optical flow $F^{'}_{k,t}$ by calculating the difference between warped coordinates $p^{'}_{t}$ and $p_{t}$. We use $F^{'}_{k,t}$ to supervise FlowNet during online adaptation:
\begin{equation}
\mathcal{L}_{flow}=\|F_{k,t}-F^{'}_{k,t}\|_{1}+\mathcal{L}_{smooth}(F_{k,t})
\end{equation}

{\bf Photometric loss} is defined in Eq.~\ref{eq:photo}. Thus the total self-supervised loss is:
\begin{equation}\label{eq:total}
\mathcal{L}=\mathcal{L}_{pho}+\mathcal{L}_{d}+\mathcal{L}_{flow}
\end{equation}


\section{Experiments}
\subsection{Implementation details}
{\bf Network Architectures} Since our method focuses on improving online adaptation of deep VO to achieve better generalization, we adopt similar networks with existing self-supervised VO methods. As for DepthNet, we use the same architecture as Monodepth2~\cite{monodepth2} and add a $5\times5$ convolution layer at the output to predict depth uncertainty map $\Sigma_{k}$. The optical flow network is based on RAFT~\cite{raft}. We add a $5\times5$ convolution + Sigmoid layer at output to predict photometric uncertainty $\Gamma_{t}$ at the same time.

\begin{figure*}
	\centering
	\subfigure[Seq.00]{\includegraphics[height=4.5cm]{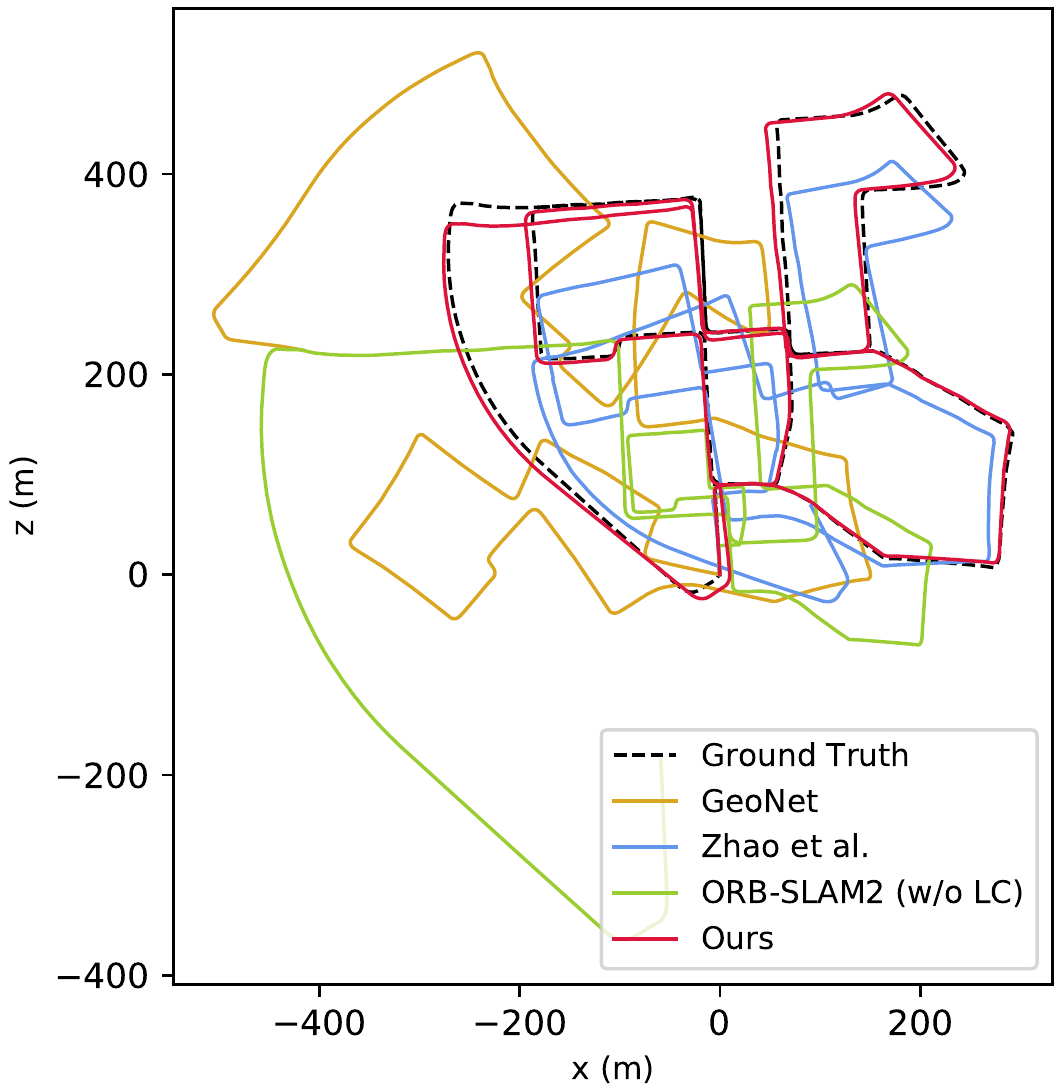}} 
	\subfigure[Seq.01]{\includegraphics[height=4.5cm]{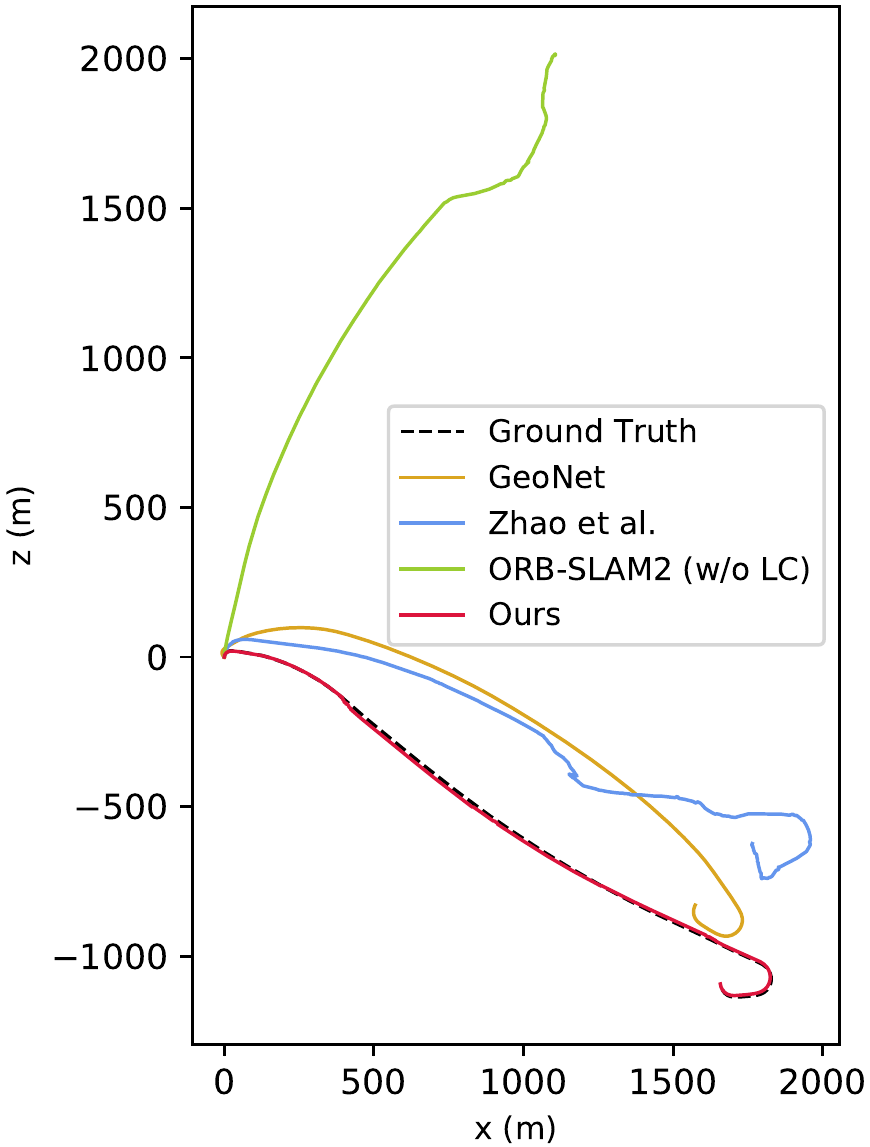}} 
	\subfigure[Seq.02]{\includegraphics[height=4.5cm]{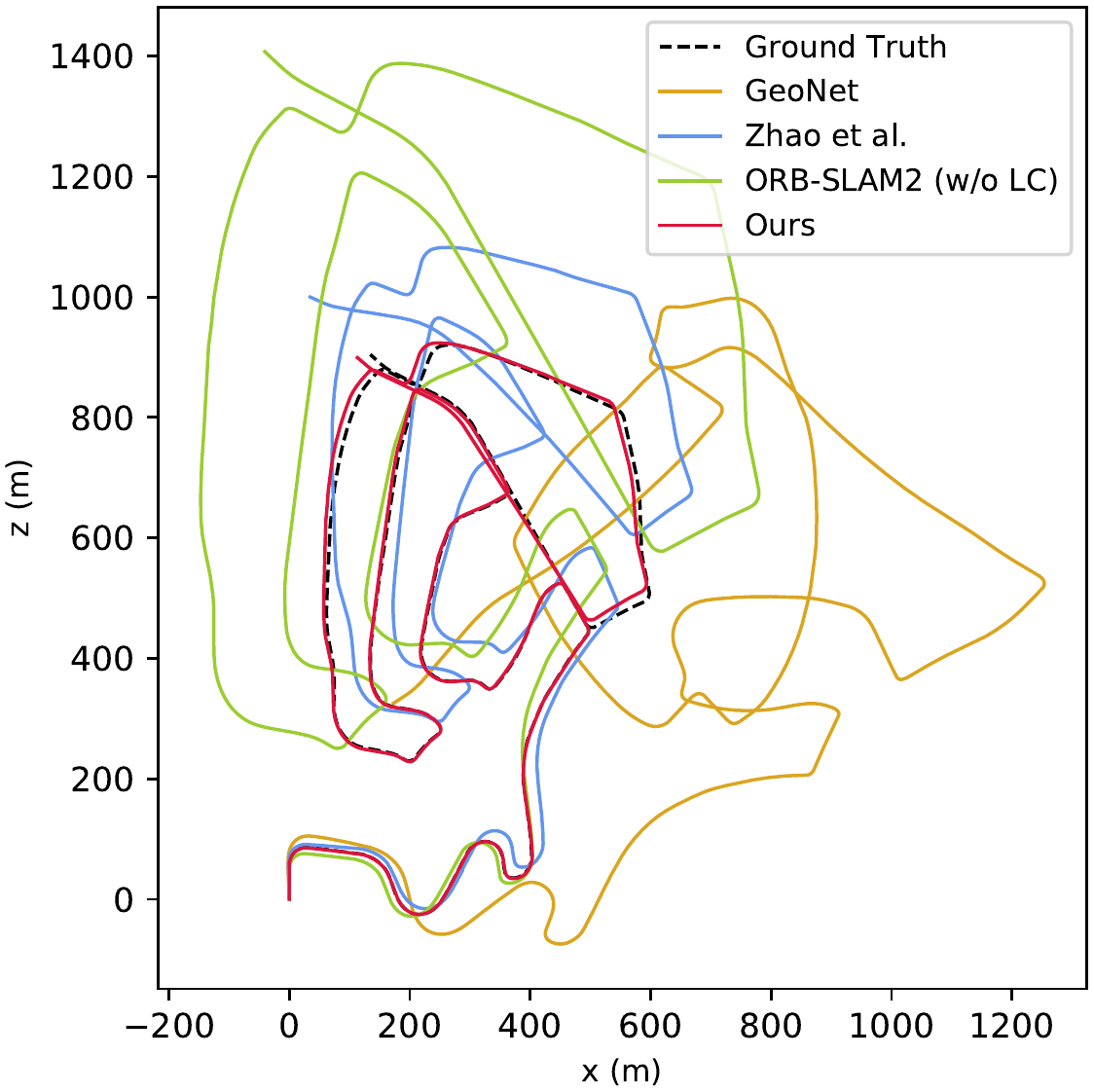}} 
	\subfigure[Seq.08]{\includegraphics[height=4.5cm]{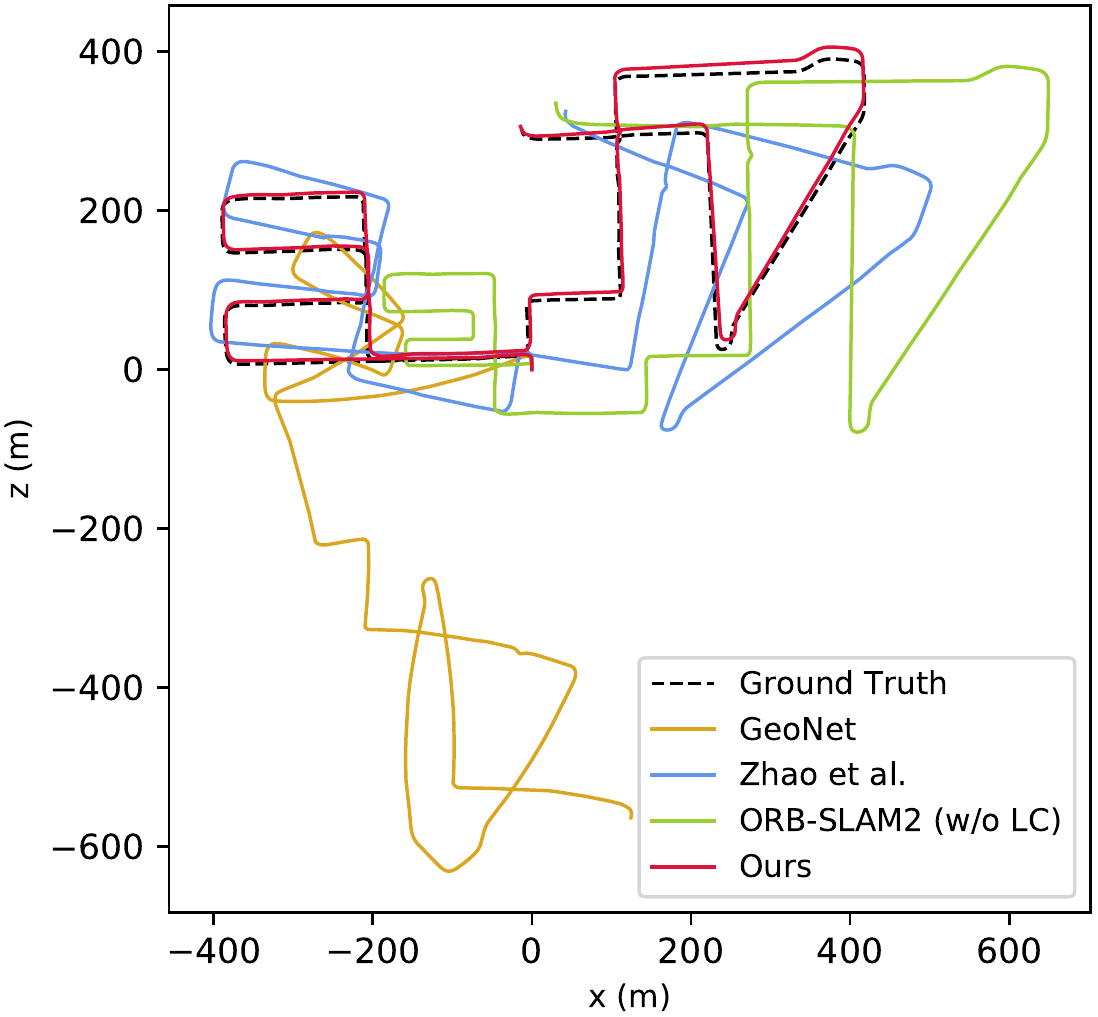}} 
	\caption{Selected trajectories of different methods on KITTI odometry dataset. We use pretrained network on Cityscapes to directly test on KITTI while all the other methods use pretrained network on KITTI for testing. It can be seen that our method shows much more accurate trajectories. Note that we do not use any mapping, pose graph optimization, loop closing or bundle adjustment techniques.}
	\label{fig:kitti_pose}
\end{figure*}

{\bf Learning Settings} Our model is implemented by PyTorch~\cite{pytorch} on a single NVIDIA GTX 2080Ti. The images are resized to $256\times832$ for KITTI~\cite{kitti} and Cityscapes~\cite{cityscapes} datasets while set $192\times256$ for TUM dataset~\cite{TUM}. The FlowNet and DepthNet are pretrained in a self-supervised manner for $1\times10^{5}$ iterations according to~\cite{competitive}. The Adam~\cite{adam} optimizer with $\beta_{1}=0.9,\beta_{2}=0.99$ is used. The learning objective (Eq.~\ref{eq:total}) is used for both pretraining and online adaptation with the learning rate of $1\times10^{-4}$. The uncertainty maps $\Gamma_{t},\Sigma_{k}$ are also jointly trained by minimizing Eq.~\ref{eq:total}. During online adaptation, we retrain FlowNet and DepthNet for 2 iterations in every time step.

\subsection{Cityscapes to KITTI}
Firstly, we try to test the generalization ability of our framework to different outdoor environments. We pretrain our method on Cityscapes~\cite{cityscapes} dataset and test on KITTI~\cite{kitti} dataset, which differ not only in scene contents and white balance but also in camera intrinsics. We compare with recent self-supervised VO baselines: GeoNet~\cite{GeoNet}, Vid2Depth~\cite{vid2depth}, Zhan~\etal~\cite{deepvofeat}, SAVO~\cite{savo} and Li~\etal~\cite{onlinevo} as well as classic methods: ORB-SLAM2~\cite{orb} (with and without loop closure) and VISO2~\cite{viso2}. Besides, we compare with Zhao~\etal~\cite{towards} and DF-VO~\cite{dfvo} which are state-of-the-art methods that combine the output of pretrained networks with classic VO pipeline.

As for pose estimation, we evaluate on 11 KITTI sequences with ground truth poses~\cite{GeoNet}. It's worthy to note that \textbf{all the other VO baselines are pretrained on KITTI}, while our method is \textbf{only pretrained on Cityscapes} and directly tested on KITTI dataset. Although in such unfair conditions, our method achieves state-of-the-art results even compared with ORB-SLAM2 (LC) (shown in Table~\ref{tab:kitti_pose} and Fig.~\ref{fig:kitti_pose}). Meanwhile, different from most self-supervised VO baselines, our method maintains a consistent scale of the entire trajectory. Thus, instead of calculating absolute trajectory error (ATE) on short sequence as previous methods, we align trajectories with ground truth~\cite{kitti} by a single scaling factor and compute translation/rotation error $t_{err}/r_{err}$ on entire trajectory.

\begin{figure}
	\centering
	\subfigure{\includegraphics[width=0.48\linewidth]{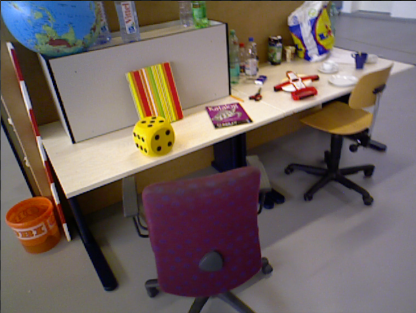}}  
	\subfigure{\includegraphics[width=0.48\linewidth]{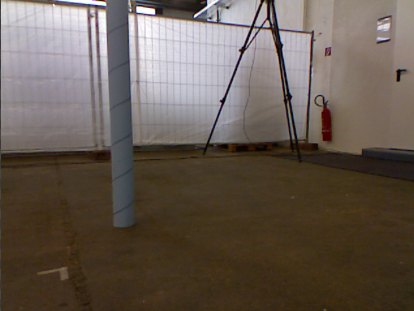}}  
	\subfigure[fr3/long\_office\_hou\_valid]{\includegraphics[width=0.48\linewidth]{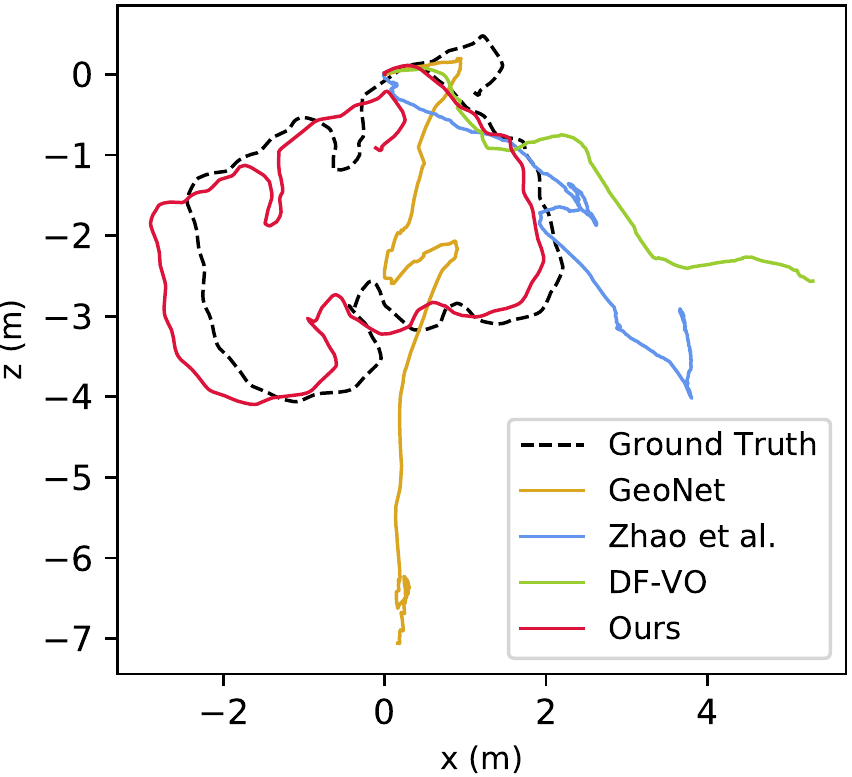}}
	\subfigure[fr2/pioneer\_360]{\includegraphics[width=0.48\linewidth]{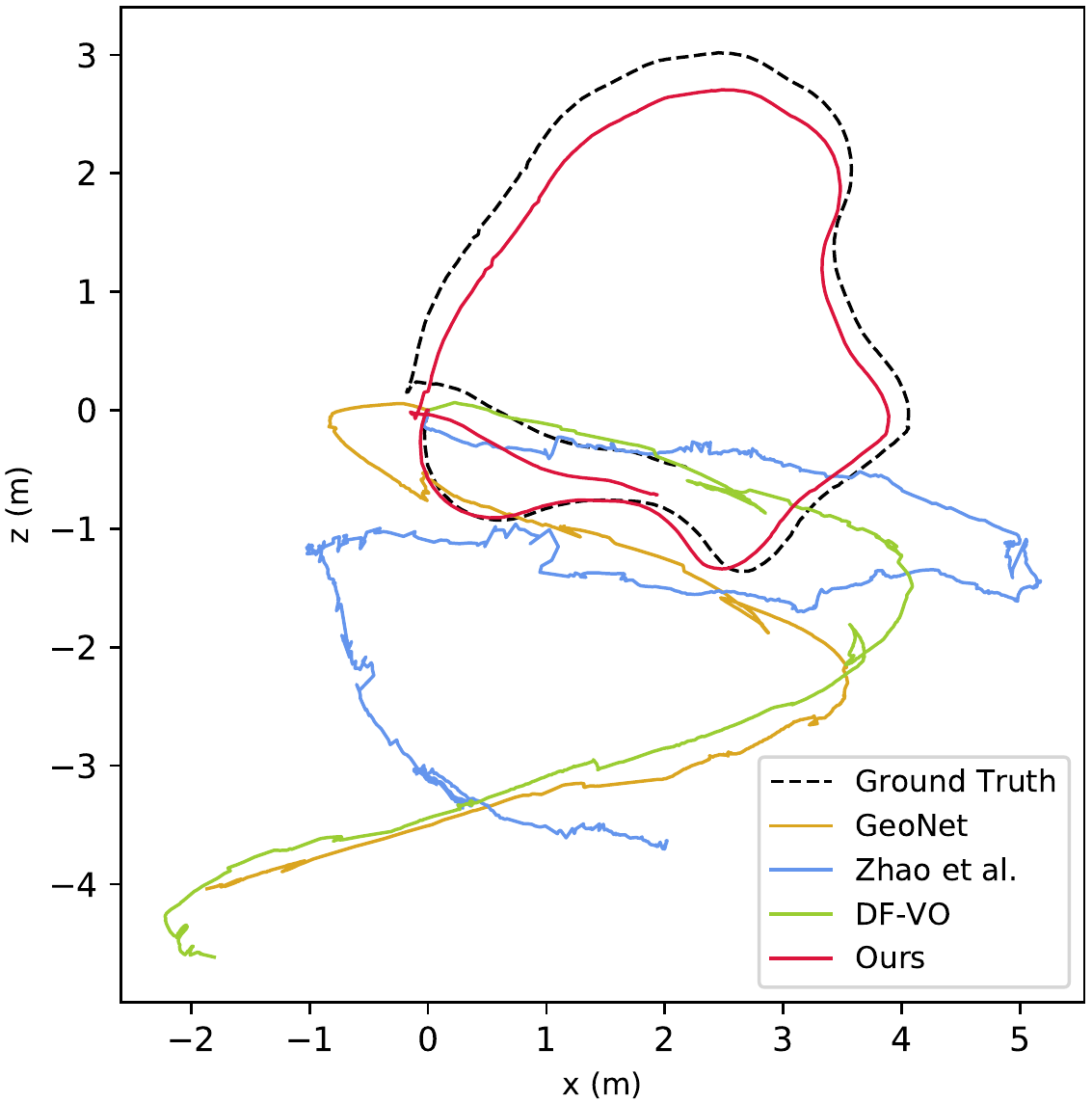}} 
	\caption{Visual odometry results pretrained on outdoor KITTI and tested on indoor TUM dataset. All the other learning-based baselines tend to fail when faced with large domain shift. In contrast, our method is still able to recover accurate VO estimation by online adapting to challenging indoor datasets.}
	\label{fig:tum_pose}
\end{figure}

Our method outperforms all the other baselines (including end-to-end learning and combination of geometric computation methods) by a clear margin. The rotation and translation errors are an order of magnitude smaller than the other self-supervised baselines, indicating that pose, depth and scale estimation collaborated with probabilistic geometric computation is much better than learning-based inference. As for classic baselines, ORB-SLAM2 is implemented by a local map tracking with bundle adjustment (BA) and ORB-SLAM2 (LC) processes the entire sequence with loop closure, pose graph optimization and global BA to ensure good performance. Our method \textbf{doesn't use any optimization backend techniques} but it still achieves comparable results with ORB-SLAM2 (LC).

\begin{table*}
	\scriptsize
	\setlength\tabcolsep{2pt}
	\renewcommand\arraystretch{1.1}
	\begin{center}
		\resizebox{\textwidth}{26.2mm}{
		\begin{tabular}{l|c|c|c|c|c|c|c|c|c|c|c}
			\hline
			\hline
			Method & Seq.00 & Seq.01 & Seq.02 & Seq.03 & Seq.04 & Seq.05 & Seq.06 & Seq.07 & Seq.08 & Seq.09 & Seq.10 \\
			& $t_{err}/r_{err}$ & $t_{err}/r_{err}$ & $t_{err}/r_{err}$ & $t_{err}/r_{err}$ & $t_{err}/r_{err}$ & $t_{err}/r_{err}$ & 
			$t_{err}/r_{err}$ & $t_{err}/r_{err}$ & $t_{err}/r_{err}$ & $t_{err}/r_{err}$ & $t_{err}/r_{err}$	\\
			\hline
			Vid2Depth~\cite{vid2depth} & 59.97~~22.59 & 9.34~~4.18 & 55.20~~14.61 & 27.02~~10.39 & 1.89~~1.19 & 51.14~~21.86 & 58.07~~26.83 & 51.21~~36.64 & 45.82~~18.10 & 44.52~~12.11 & 21.45~~12.50 \\
			GeoNet~\cite{GeoNet} & 27.60~~5.72 & 12.25~~4.15 & 42.21~~6.14 & 19.21~~9.78 & 9.09~~7.55 & 20.12~~7.67 & 9.28~~4.34 & 8.27~~5.93 & 18.59~~7.85 & 23.94~~9.81 & 20.73~~9.10 \\
			Zhan \textit{et al.}~\cite{deepvofeat} & 6.23~~2.44 & 23.78~~1.75 & 6.59~~2.26 & 15.76~~10.62 & 3.14~~2.02 & 4.94~~2.34 & 5.80~~2.06 & 6.49~~3.56 & 5.45~~2.39 & 11.89~~3.62 & 12.82~~3.40 \\
			SAVO~\cite{savo} & 18.67~~3.12 & 9.86~~1.23 & 17.58~~4.29 & 15.01~~6.54 & 3.35~~1.18 & 9.82~~2.53 & 5.27~~4.30 & 9.85~~4.03 & 21.37~~3.65 & 9.52~~3.64 & 6.45~~2.41  \\
			Li~\etal~\cite{onlinevo} & 8.42~~3.91 & 17.36~~4.60 & 14.38~~2.62 & 18.24~~0.92 & 3.28~~4.40 & 7.58~~3.31 & 4.36~~2.28 & 5.58~~3.12 & 7.51~~2.63 & 5.89~~3.34 & 4.79~~0.83 \\
			\hline
			VISO2~\cite{viso2} & 12.66~~2.73 & 41.93~~7.68 & 9.47~~1.19 & 3.93~~2.21 & 2.50~~1.78 & 15.10~~3.65 & 6.80~~1.93 & 10.80~~4.67 & 14.82~~2.52 & 3.69~~1.25 & 21.01~3.26 \\
			DF-VO~\cite{dfvo} & 2.25~~0.58 & 66.98~~17.04 & 3.60~~0.52 & 2.67~~0.50 & 1.43~~0.29 & \underline{1.10}~~0.30 & 1.03~~0.30 & \textbf{0.97}~~\textbf{0.27} & 1.60~~0.32 & 2.61~~0.29 & 2.29~~0.37 \\
			D3VO~\cite{d3vo} (stereo) & -~~~~~~- & \textbf{1.07}~~~~~- & \textbf{0.80}~~~~~- & -~~~~~~- & -~~~~~~- & -~~~~~~- & \textbf{0.67}~~~~~- & -~~~~~~- & \textbf{1.00}~~~~~- & \textbf{0.78}~~~~~- & \textbf{0.62}~~~~~- \\
			Zhao~\etal~\cite{towards} & 4.45~~1.13 & 62.54~~2.71 & 4.64~~0.91 & 6.86~~1.26 & 4.76~~3.31 & 2.93~~0.90 & 3.48~~1.32 & 2.57~~1.21 & 5.09~~1.19 & 6.81~~0.72 & 4.39~~1.05  \\
			ORB-SLAM2~\cite{orb} & 11.43~~0.58 & 107.57~~0.89 & 10.34~~\textbf{0.26} & \underline{0.97}~~\textbf{0.19} & 1.30~~0.27 & 9.04~~\underline{0.26} & 14.56~~\underline{0.26} & 9.77~~0.36 & 11.46~\textbf{0.28} & 9.30~~\underline{0.26} & 2.57~~\underline{0.32} \\
			ORB-SLAM2 (LC) & 2.35~~\textbf{0.35} & 109.10~~\textbf{0.45} & 3.32~~\underline{0.31} & \textbf{0.91}~~\textbf{0.19} & 1.56~~0.27 & 1.84~~\textbf{0.20} & 4.99~~\textbf{0.23} & \underline{1.91}~\underline{0.28} & 9.41~~\underline{0.30} & 2.88~~\textbf{0.25} & 3.30~~\textbf{0.30} \\
			\hline
			\textbf{Ours (w/o RDS)} & 4.67~~1.28 & 6.99~~2.83 & 4.33~~1.05 & 8.73~~1.14 & 3.78~~2.09 & 4.20~~1.98 & 5.02~~3.61 & 7.24~~1.11 & 3.30~~2.78 & 7.99~~2.53 & 5.21~~2.87 \\
			\textbf{Ours (w/o PU)} & 2.28~~0.87 & 5.42~~1.40 & 3.98~~1.87 & 7.76~~0.99 & 2.92~~1.04 & 3.63~~1.28 & 4.92~~2.07 & 8.25~~2.39 & 3.28~~1.69 & 4.60~~1.13 & 3.25~~1.70  \\
			\textbf{Ours} & \textbf{1.32}~~\underline{0.45} & \underline{2.83}~~\underline{0.65} & \underline{1.42}~~0.45 & 1.77~~\underline{0.39} & \textbf{1.22}~~\textbf{0.27} & \textbf{1.07}~~0.44 & \underline{1.02}~~0.41 & 2.06~~1.18 & \underline{1.50}~~0.42 & \underline{1.87}~~0.46 & \underline{1.93}~~\textbf{0.30} \\
			\hline
			\hline
		\end{tabular}}
	\end{center}

	\caption{Quantitative comparison on KITTI dataset. Our method is pretrained on Cityscapes and tested on KITTI, while \textbf{all the other learning-based methods are pretrained on KITTI}. LC: loop closure, w/o: without, RDS: refined depth for online supervision, PU: photometric uncertainty. $t_{err}$: translational root mean square error (RMSE) drift (\%); $r_{err}$: average rotational RMSE drift (${}^{\circ}$/100m).}
	\label{tab:kitti_pose}
\end{table*}


\begin{table*}
	\footnotesize
	\setlength\tabcolsep{2.4pt}
	\begin{center}
		\begin{tabular}{l|ccccc|cccc|ccc}
			\hline
			\hline
			& Vid2Depth & GeoNet & Zhan \etal & SAVO & Li \etal & DF-VO & Zhao \etal & DSO & ORB-SLAM2 & Ours & Ours & Ours \\
			Sequence & \cite{vid2depth} & \cite{GeoNet} & \cite{deepvofeat} & \cite{savo} & \cite{onlinevo} &
			\cite{dfvo} & \cite{towards} & \cite{DSO} & (LC)~\cite{orb} & & (w/o RDS) & (w/o PU) \\
			\hline
			fr2/desk & 0.698 & 0.462 & 0.570 & 0.402 & 0.214 & 0.306 & 0.485 & X & X & \textbf{0.158} & 0.572 & 0.221 \\ 
			fr2/pioneer\_360 & 0.581 & 0.662 & 0.453 & 0.402 & 0.218 & 0.599 & 0.693 & X & X & \textbf{0.201} & 0.638 & 0.254 \\ 
			fr2/pioneer\_slam & 0.367 & 0.301 & 0.309 & 0.338 & 0.190 & 0.585 & 0.354 & 0.737 & X & \textbf{0.176} & 0.481 & 0.210 \\ 
			fr2/360\_kidnap & 0.564 & 0.579 & 0.430 & 0.421 & 0.357 & 0.745 & 0.468 & X & 0.582 & 0.384 & 0.605 & \textbf{0.371} \\ 
			fr3/cabinet 	& 0.492 & 0.282 & 0.316 & 0.281 & 0.272 & 0.447 & 0.227 & X & X & \textbf{0.213} & 0.453 & 0.276 \\     
			fr3/long\_office\_hou\_valid & 0.401 & 0.316 & 0.327 & 0.297 & 0.237 & 0.227 & 0.534 & 0.327 & \textbf{0.042} & 0.133 & 0.529 & 0.168 \\ 
			fr3/nostr\_texture\_near\_loop & 0.328 & 0.277 & 0.340 & 0.440 & 0.255 & 0.564 & 0.348 & 0.093 & \textbf{0.057} & 0.159 & 0.401 & 0.186 \\ 
			fr3/str\_notexture\_far & 0.227 & 0.258 & 0.235 & 0.216 & 0.177 & 0.505 & 0.175 & 0.543 & X & \textbf{0.104} & 0.432 & 0.201 \\ 
			fr3/str\_notexture\_near & 0.235 & 0.198 & 0.217 & 0.204 & 0.128 & 0.603 & 0.218 & 0.481 & X & \textbf{0.207} & 0.579 & 0.224 \\ 
			\hline
			\hline
		\end{tabular}
	\end{center}
	\caption{Quantitative evaluation of different methods pretrained on KITTI and tested on TUM-RGBD dataset. We evaluate relative pose error (RPE) which is presented as translational RMSE in [m/s]. LC: loop closure, X: fail. w/o RDS: without refined depth for online supervision. w/o PU: without online learned photometric uncertainty.}
	\label{tab:TUM_pose}
\end{table*}

\subsection{Outdoor KITTI to indoor TUM}

In order to further evaluate the generalization ability to more complex indoor environments, we test on TUM~\cite{TUM} dataset using networks pretrained on KITTI. TUM indoor dataset contains much more complicated motion patterns and challenging conditions. As shown in Table~\ref{tab:TUM_pose} and Fig.~\ref{fig:tum_pose}, learning-based baselines have large errors when confronted with significant domamin shift and different motion patterns (from fast planar motion to small motion in $xyz$ axies). On the contrary, our method yields promising results due to fast online adaptation. Besides, our method is more robust than classic methods (ORB-SLAM2~\cite{orb} and DSO~\cite{DSO}) in textureless scenes, abrupt motion and illumination changes, indicating that it tends to find out robust correspondences and online learns depth/photometric uncertainty in challenging conditions.


\subsection{Depth evaluation on KITTI and NYUv2}
We demonstrate the effectiveness of using optimized $D^{'}_{k}$ for self-supervision by evaluating different single-view depth estimation methods on KITTI~\cite{kitti} and NYUv2~\cite{nyu} datasets. We only use triangulation and Bayesian updating for training. During test, our method predicts \textit{single-view depth} without refinement. As for KITTI, we take Eigen \etal~\cite{eigen} split for training and test. As for NYUv2, we use the raw training set and evaluate depth prediction results on labeled test set. The predicted depth is multiplied by a scaling factor to match the median with ground truth~\cite{eigen}.

\begin{figure*}
	\centering
	\includegraphics[width=0.99\linewidth]{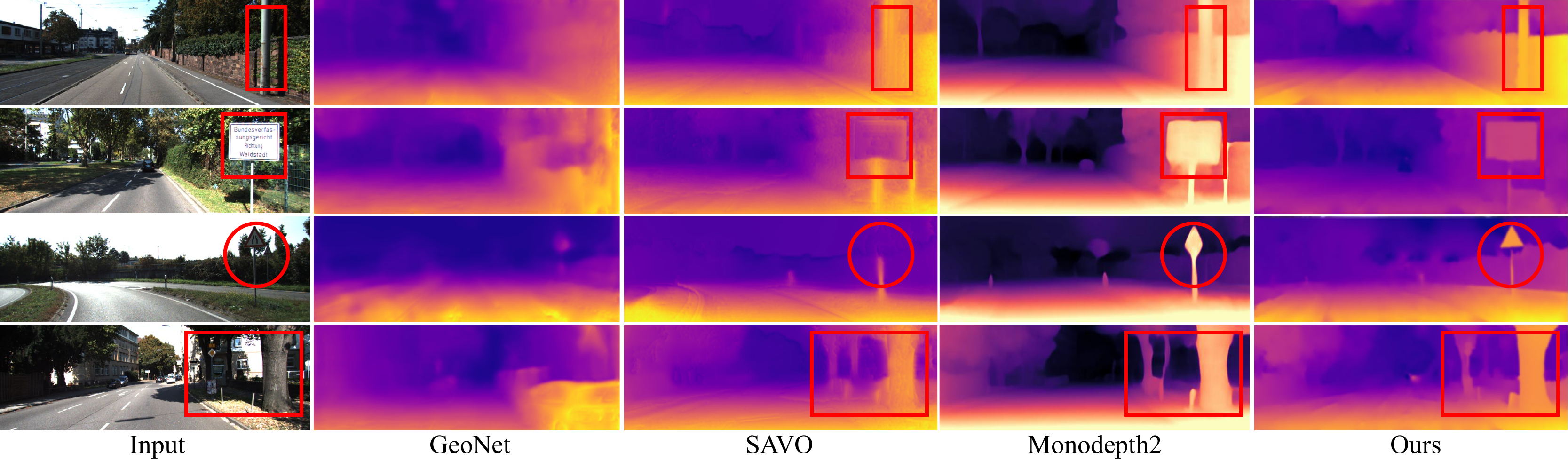}
	\caption{Depth estimation results on KITTI dataset. Thanks to our triangulation process and multi-frame depth refinement, our method shows better predictions and preserves sharp edges while other methods tend to predict vague depth. Best viewed in color.}
	\label{fig:depth_kitti}
\end{figure*}

Table~\ref{tab:kitti_depth},~\ref{tab:nyu_depth} and Fig.~\ref{fig:depth_kitti} show the depth evaluation results on KITTI and NYUv2 datasets. Benefiting from the patch-based depth triangulation and multi-frame refinement process, our method is able to synthesize refined depth for self-supervision. The learned depth is more accurate and preserves sharper edges with fine details than other methods. More qualitative results and analysis can be found in the supplementary materials.

\begin{table*}
	\small
	\begin{center}
		\begin{tabular}{l|c|c|c|c|c|c|c|c}
			\hline
			\hline
			Method & Supervision & Abs Rel & Sq Rel & RMSE & RMSE log & $\delta<1.25$ & $\delta<1.25^{2}$ & $\delta<1.25^{3}$ \\
			\hline
			SfMLearner~\cite{SfMLearner} & - & 0.208   & 1.768  & 6.856 & 0.283 & 0.678 & 0.885 & 0.957 \\
			Garg~\etal~\cite{garg} & stereo & 0.169   & 1.080  & 5.104 & 0.273 & 0.740 & 0.904 & 0.962 \\
			Vid2Depth~\cite{vid2depth} & - & 0.163   & 1.240  & 6.220 & 0.250 & 0.762 & 0.916 & 0.968 \\
			GeoNet~\cite{GeoNet} & - & 0.155   & 1.296  & 5.857 & 0.233 & 0.793 & 0.931 & 0.973 \\
			Zhan~\etal~\cite{deepvofeat} & stereo & 0.135 & 1.132 & 5.585 & 0.229 & 0.820 & 0.933 & 0.971 \\
			Mahjourian~\etal~\cite{mah} & - & 0.163 & 1.240 & 6.220 & 0.250 & 0.762 & 0.916 & 0.968 \\
			SAVO~\cite{savo} & - & 0.150 & 1.127 & 5.564 & 0.229 & 0.823 & 0.936 & 0.974  \\
			SC-SfMLearner~\cite{scsfm} & - & 0.137 & 1.089 & 5.439 & 0.217 & 0.830 & 0.942 & 0.975 \\
			Zhao~\etal~\cite{towards} & - & 0.113 & 0.704 & 4.581 & \textbf{0.184} & 0.871 & 0.961 & 0.984 \\
			Monodepth2~\cite{monodepth2} (w/o pretrain) & - & 0.132 & 1.044 & 5.142 & 0.210 & 0.845 & 0.948 & 0.977 \\
			Monodepth2 (ImageNet pretrain) & - & 0.115 & 0.882 & 4.701 & 0.190 & 0.879 & 0.961 & 0.982 \\
			Ranjan~\etal~\cite{competitive} & - & 0.148 & 1.149 & 5.464 & 0.226 & 0.815 & 0.935 & 0.973 \\
			\hline
			\textbf{Ours (w/o RDS)} & - & 0.136 & 1.087 & 5.118 & 0.210 & 0.843 & 0.952 & 0.980 \\
			\textbf{Ours (w/o PU)} & - & 0.115 & 0.799 & 4.282 & 0.253 & 0.882 & 0.965 & 0.981 \\
			\textbf{Ours} & - & \textbf{0.106} & \textbf{0.701} & \textbf{4.129} & 0.210 & \textbf{0.889} & \textbf{0.967} & \textbf{0.984} \\
			\hline
			\hline
		\end{tabular}
	\end{center}
	\caption{Depth estimation results on KITTI dataset by Eigen \textit{et al.}~\cite{eigen} split. The results are capped at 80 meters. As for error metrics Abs Rel, Seq Rel, RMSE and RMSE log, lower value is better; as for accuracy metrics $\delta$, higher value is better. w/o RDS: without refined depth for online supervision. w/o PU: without online learned photometric uncertainty.}
	\label{tab:kitti_depth}
\end{table*}

\begin{table}
	\footnotesize
	\setlength\tabcolsep{3.7pt}
	\begin{center}
		\begin{tabular}{l|ccc|ccc}
			\hline
			\hline
			& & Error & &\multicolumn{3}{c}{Accuracy $\delta<1.25^{n}$} \\
			Method & Rel & log10 & RMSE & $n=1$ & $n=2$ & $n=3$ \\
			\hline
			Make3D~\cite{make3d} & 0.349 & - & 1.214 & 0.447 & 0.745 & 0.987 \\
			Li \etal~\cite{libo} & 0.232 & 0.094 & 0.821 & 0.621 & 0.886 & 0.968 \\
			MS-CRF~\cite{mscrf} & 0.121 & 0.052 & 0.586 & 0.811 & 0.954 & 0.987 \\
			DORN~\cite{dorn} & \underline{0.115} & \underline{0.051} & \underline{0.509} & \underline{0.828} & 0.965 & \underline{0.992} \\
			\hline
			Zhou~\etal~\cite{moving} & 0.208 & 0.086 & 0.712 & 0.674 & 0.900 & 0.968 \\
			Zhao~\etal~\cite{towards} & 0.201 & 0.085 & 0.708 & 0.687 & 0.903 & 0.968 \\
			$\text{P}^{2}$Net*~\cite{p2net} & 0.147 & \textbf{0.062} & 0.553 & 0.801 & 0.951 & 0.987 \\
			\hline
			\textbf{Ours (w/o RDS)} & 0.225 & 0.090 & 0.702 & 0.711 & 0.882 & 0.970 \\
			\textbf{Ours (w/o PU)} & 0.142 & 0.087 & 0.631 & 0.784 & 0.923 & 0.976 \\
			\textbf{Ours} & \textbf{0.139} & 0.071 & \textbf{0.528} & \textbf{0.805} & \textbf{0.967} & \textbf{0.989} \\
			\hline
			\hline
		\end{tabular}
	\end{center}
	\caption{Depth estimation results on NYUv2 dataset. Supervised methods are shown in the first rows. *The inference resolution of $\text{P}^{2}$Net is $288\times384$ with 5-frame left-right flipping augmentation.}
	\label{tab:nyu_depth}
\end{table}

\subsection{Ablation studies}
In order to demonstrate the effectiveness of each component, we present ablation studies on various versions of our method on KITTI, TUM and NYUv2 datasets (shown in Table~\ref{tab:kitti_pose},~\ref{tab:TUM_pose},~\ref{tab:kitti_depth},~\ref{tab:nyu_depth}). `w/o RDS' means without the final step of retraining both DepthNet and FlowNet. It can be seen that the performance of pose and depth estimation shows a considerable improvement when the refined depth is used for online training the DepthNet. Besides, it can be noticed that KITTI contains many moving objects (cars, people) and all these datasets have many sequences with changing camera exposure time. The online learned photometric uncertainty (w/o PU) helps a lot on KITTI and TUM for pose estimation. We suggest readers to refer to supplementary materials for more qualitative comparisons.


\section{Conclusions}
In this paper, we propose an online adaptation framework for deep VO with the assistance of scene-agnostic geometric computations and Bayesian inference. The predicted single-view depth is continuously improved with incoming observations by Baysian depth filter. Meanwhile, we explicitly model depth and photometric uncertainties to deal with the observation noise. The optimized pose, depth and flow from differentiable Gauss-Newton layer are used for online self-supervision. Extensive experiments on various environment shifting demonstrate that our method has much better generalization ability than state-of-the-art learning-based VO methods.

\textbf{Acknowledgments}
This work is supported by the National Key Research and Development Program of China (2017YFB1002601) and National Natural Science Foundation of China (61632003, 61771026).

{\small
\bibliographystyle{ieee_fullname}
\bibliography{egbib}
}

\end{document}